\documentclass[11pt,a4paper]{article}
\usepackage[hyperref]{acl2020}
\usepackage{times}
\usepackage{latexsym}
\usepackage{graphicx}
\usepackage{amsmath}
\usepackage{amssymb}
\usepackage{booktabs}
\usepackage{multirow}
\usepackage{relsize}
\usepackage{caption}
\usepackage{subcaption}
\usepackage{soul}

\usepackage{microtype}

\aclfinalcopy 
\def\aclpaperid{2474} 

\newcommand{\orange}[1]{{\color[HTML]{ff7f0e}\textbf{#1}}}
\newcommand{\blue}[1]{{\color[HTML]{1f77b4}\textbf{#1}}}
\newcommand{\green}[1]{{\color[HTML]{2ca02c}\textbf{#1}}}
\newcommand{\gray}[1]{{\color[HTML]{808080}\textbf{#1}}}

\title{Fact-based Text Editing}

\author{
    Hayate Iso\textsuperscript{\dag\thanks{~~The work was done when Hayate Iso was a research intern at ByteDance AI Lab.}} \quad Chao Qiao\textsuperscript{\ddag} \quad Hang Li\textsuperscript{\ddag}\\
    \textsuperscript{\dag}Nara Institute of Science and Technology~
    \textsuperscript{\ddag}ByteDance AI Lab\\
    \texttt{hyate.iso@gmail.com}, 
    \texttt{\{qiaochao, lihang.lh\}@bytedance.com}
}

\date{}

\begin{document}
\maketitle
\begin{abstract}
We propose a novel text editing task, referred to as \textit{fact-based text editing}, in which the goal is to revise a given document to better describe the facts in a knowledge base (e.g., several triples). The task is important in practice because reflecting the truth is a common requirement in text editing. First, we propose a method for automatically generating a dataset for research on fact-based text editing, where each instance consists of a draft text, a revised text, and several facts represented in triples. We apply the method into two public table-to-text datasets, obtaining two new datasets consisting of 233k and 37k instances, respectively. Next, we propose a new neural network architecture for fact-based text editing, called \textsc{FactEditor}, which edits a draft text by referring to given facts using a buffer, a stream, and a memory. A straightforward approach to address the problem would be to employ an encoder-decoder model. Our experimental results on the two datasets show that \textsc{FactEditor} outperforms the encoder-decoder approach in terms of fidelity and fluency. The results also show that \textsc{FactEditor} conducts inference faster than the encoder-decoder approach. 
\end{abstract}

\section{Introduction}
\begin{table}[t]
    \centering
    \small
    \begin{tabular}{l}
        \toprule
        \textbf{Set of triples} \\
        \begin{tabular}{@{}c@{}l@{\;}l@{\;}l@{}}
            \{ & (\green{Baymax}, & \green{creator}, & \green{Douncan\_Rouleau}),\\
             & (\blue{Douncan\_Rouleau}, & \blue{nationality}, & \blue{American}),\\
             & (\blue{Baymax}, & \blue{creator}, & \blue{Steven\_T.\_Seagle}),\\
              & (\blue{Steven\_T.\_Seagle}, & \blue{nationality}, & \blue{American}),\\
             & (\green{Baymax}, & \green{series},& \green{Big\_Hero\_6}),\\
             & (\blue{Big\_Hero\_6}, & \blue{starring}, &\blue{Scott\_Adsit})\}
        \end{tabular}\\
        \midrule
        \textbf{Draft text}\\
        \green{Baymax} was created by \green{Duncan\_Rouleau}, \orange{a winner of}\\
        \orange{Eagle\_Award}.
        \green{Baymax} is a character in \green{Big\_Hero\_6} .\\
        \midrule
        \textbf{Revised text}\\
        \green{Baymax} was created by \blue{American} creators\\
        \green{Duncan\_Rouleau} and \blue{Steven\_T.\_Seagle} . \green{Baymax} is\\
        a character in \green{Big\_Hero\_6} which stars \blue{Scott\_Adsit} .\\
        \bottomrule
    \end{tabular}
    \caption{Example of fact-based text editing. Facts are represented in triples. The \green{facts in green} appear in both draft text and triples.  The \orange{facts in orange} are present in the draft text, but absent from the triples. The \blue{facts in blue} do not appear in the draft text, but in the triples.  The task of \textit{fact-based text editing} is to edit the draft text on the basis of the triples, by deleting \orange{unsupported facts} and inserting \blue{missing facts} while retaining \green{supported facts}.}
    \label{tab:problem}
\end{table}

Automatic editing of text by computer is an important application, which can help human writers to write better documents in terms of accuracy, fluency, etc. The task is easier and more practical than the automatic generation of texts from scratch and is attracting attention recently~\cite{yang2017identifying,yin2019learning}. In this paper, we consider a new and specific setting of it, referred to as {\em fact-based text editing}, in which a draft text and several facts (represented in triples) are given, and the system aims to revise the text by adding missing facts and deleting unsupported facts. Table~\ref{tab:problem} gives an example of the task.

As far as we know, no previous work did address the problem. In a text-to-text generation, given a text, the system automatically creates another text, where the new text can be a text in another language (machine translation), a summary of the original text (summarization), or a text in better form (text editing). In a table-to-text generation, given a table containing facts in triples, the system automatically composes a text, which describes the facts. The former is a text-to-text problem, and the latter a table-to-text problem. In comparison, fact-based text editing can be viewed as a `text\&table-to-text' problem.

First, we devise a method for automatically creating a dataset for fact-based text editing. 
Recently, several table-to-text datasets have been created and released, consisting of pairs of facts and corresponding descriptions.
We leverage such kind of data in our method. We first retrieve facts and their descriptions. Next, we take the descriptions as revised texts and automatically generate draft texts based on the facts using several rules. We build two datasets for fact-based text editing on the basis of \textsc{WebNLG}~\cite{gardent-etal-2017-creating} and \textsc{RotoWire}, consisting of 233k and 37k instances respectively~\cite{wiseman2017challenges}~\footnote{The datasets are publicly available at \url{https://github.com/isomap/factedit}}.

Second, we propose a model for fact-based text editing called \textsc{FactEditor}. One could employ an encoder-decoder model, such as an encoder-decoder model, to perform the task.
The encoder-decoder model implicitly represents the actions for transforming the draft text into a revised text. In contrast, \textsc{FactEditor} explicitly represents the actions for text editing, including \texttt{Keep}, \texttt{Drop}, and \texttt{Gen}, which means retention, deletion, and generation of word respectively. The model utilizes a buffer for storing the draft text, a stream to store the revised text, and a memory for storing the facts.  It also employs a neural network to control the entire editing process. \textsc{FactEditor} has a lower time complexity than the encoder-decoder model, and thus it can edit a text more efficiently.

Experimental results show that \textsc{FactEditor} outperforms the baseline model of using encoder-decoder for text editing in terms of fidelity and fluency, and also show that \textsc{FactEditor} can perform text editing faster than the encoder-decoder model.

\section{Related Work}
\subsection{Text Editing}
Text editing has been studied in different settings such as automatic post-editing~\cite{knight1994automated,simard2007statistical,yang2017identifying}, paraphrasing~\cite{dolan-brockett-2005-automatically}, sentence simplification~\cite{inui-etal-2003-text, wubben-etal-2012-sentence}, grammar error correction~\cite{ng2014conll}, and text style transfer~\cite{shen2017style,hu2017toward}. 

The rise of encoder-decoder models~\cite{cho2014learning,sutskever2014sequence} as well as the attention~\cite{bahdanau2015neural,vaswani2017attention} and copy mechanisms~\cite{gu-etal-2016-incorporating,gulcehre-etal-2016-pointing} has dramatically changed the landscape, and now one can perform the task relatively easily with an encoder-decoder model such as Transformer provided that a sufficient amount of data is available. For example, \citet{li2018paraphrase} introduce a deep reinforcement learning framework for paraphrasing, consisting of a generator and an evaluator. \citet{yin2019learning} formalize the problem of text edit as learning and utilization of edit representations and propose an encoder-decoder model for the task. \citet{zhao2018integrating} integrate paraphrasing rules with the Transformer model for text simplification. \citet{zhao2019improving} proposes a method for English grammar correction using a Transformer and copy mechanism. 

Another approach to text editing is to view the problem as sequential tagging instead of encoder-decoder. In this way, the efficiency of learning and prediction can be significantly enhanced. \citet{vu2018automatic} and \citet{dong2019editnts} conduct automatic post-editing and text simplification on the basis of edit operations and employ Neural Programmer-Interpreter~\cite{reed2016neural} to predict the sequence of edits given a sequence of words, where the edits include \texttt{KEEP}, \texttt{DROP}, and \texttt{ADD}. \citet{malmi2019lasertagger} propose a sequential tagging model that assigns a tag (\texttt{KEEP} or \texttt{DELETE}) to each word in the input sequence and also decides whether to add a phrase before the word. Our proposed approach is also based on sequential tagging of actions. It is designed for fact-based text editing, not text-to-text generation, however.

\begin{table*}[t]
    \centering
    \scriptsize
    \begin{subtable}[t]{\textwidth}
        \begin{tabular*}{\textwidth}{cc@{\quad}c@{\quad}c@{\quad}c@{\quad}c@{\quad}c@{\quad}c@{\quad}c@{\quad}c@{\quad}c@{\quad}c@{\quad}c@{\quad}c@{\quad}c@{\quad}c@{\quad}c@{\quad}c@{\quad}c@{\quad}c@{}}
            $\boldsymbol{y}'$     &  \green{AGENT-1} & \blue{performed} & \green{as} & \green{PATIENT-3} & \blue{on} & & & & & & & \green{BRIDGE-1} & \green{mission} & \orange{that} & \orange{was} & \orange{operated} & \orange{by} & \orange{PATIENT-2} & \green{.}\\
            $\hat{\boldsymbol{x}}'$ &  \green{AGENT-1} & \gray{served} & \green{as} & \green{PATIENT-3} & \multicolumn{7}{l}{\gray{was a crew member of the}} & \green{BRIDGE-1} & \green{mission} & & & &  & & \green{.}\\
            $\boldsymbol{x}'$       &  \green{AGENT-1} & \blue{performed} & \green{as} & \green{PATIENT-3} & \blue{on} & & & & & & & \green{BRIDGE-1} & \green{mission} & & & & & & \green{.}
        \end{tabular*}
        \caption{Example for insertion. The revised template $\boldsymbol{y}'$ and the reference template $\hat{\boldsymbol{x}}'$ share subsequences. The set of triple templates $\mathcal{T}\backslash\hat{\mathcal{T}}$ is \{(BRIDGE-1, operator, PATIENT-2)\}. Our method removes ``that was operated by PATIENT-2'' from the revised template $\boldsymbol{y}'$ to create the draft template $\boldsymbol{x}'$.}
        \label{tab:insert}
    \end{subtable}
    \newline
    \vspace{1em}
    \newline
    \begin{subtable}[t]{\textwidth}
        \begin{tabular*}{\textwidth}{cc@{\quad}c@{\quad}c@{\quad}c@{\quad}c@{\quad}c@{\quad}c@{\quad}c@{\quad}c@{\quad}c@{\quad}c@{\quad}c@{\quad}c@{\quad}c@{\quad}c@{\quad}c@{\quad}c@{\quad}c@{}}
            $\boldsymbol{y}'$     & & & & \green{AGENT-1} & & & & & & & & \green{was} & \green{created} & \green{by} & \green{BRIDGE-1} & \green{and} & \green{PATIENT-2} & \green{.}\\
            $\hat{\boldsymbol{x}}'$ & \multicolumn{3}{c}{\gray{The character of}} & \green{AGENT-1} & \blue{,} & \blue{whose} & \blue{full} & \blue{name} & \blue{is} & \blue{PATIENT-1} & \blue{,} & \green{was} & \green{created} & \green{by} & \green{BRIDGE-1} & \green{and} & \green{PATIENT-2} & \green{.}\\
            $\boldsymbol{x}'$     & & & & \green{AGENT-1} & \blue{,} & \blue{whose} & \blue{full} & \blue{name} & \blue{is} & \blue{PATIENT-1} & \blue{,} & \green{was} & \green{created} & \green{by} & \green{BRIDGE-1} & \green{and} & \green{PATIENT-2} & \green{.}\\
        \end{tabular*}
        \caption{Example for deletion. The revised template $\boldsymbol{y}'$ and the reference template $\hat{\boldsymbol{x}}'$ share subsequences. The set of triple templates $\hat{\mathcal{T}}\backslash\mathcal{T}$ is \{(AGENT-1, fullName, PATIENT-1)\}. Our method copies ``whose full name is PATIENT-1'' from the reference template $\boldsymbol{x}'$ to create the draft template $\boldsymbol{x}'$.}
        \label{tab:delete}
    \end{subtable}
    \caption{Examples for insertion and deletion, where \green{words in green} are matched, \gray{words in gray} are not matched, \blue{words in blue are copied}, and \orange{words in orange} are removed. Best viewed in color.}
    \label{tab:edit}
\end{table*}
\subsection{Table-to-Text Generation}
Table-to-text generation is the task which aims to generate a text from structured data~\cite{reiter2000building,gatt2018survey}, for example, a text from an infobox about a term in biology in wikipedia~\cite{lebret2016neural} and a description of restaurant from a structured representation~\cite{novikova2017e2e}. Encoder-decoder models can also be employed in table-to-text generation with structured data as input and generated text as output, for example, as in ~\cite{lebret2016neural}.
\citet{puduppully2019data} and \citet{iso2019learning} propose utilizing an entity tracking module for document-level table-to-text generation.

One issue with table-to-text is that the style of generated texts can be diverse~\cite{iso2019learning}. Researchers have developed methods to deal with the problem using other texts as templates~\cite{hashimoto2018retrieve,guu2018generating,peng2019text}. The difference between the approach and fact-based text editing is that the former is about table-to-text generation based on other texts, while the latter is about text-to-text generation based on structured data.

\section{Data Creation}
\label{sec:make_data}

In this section, we describe our method of data creation for fact-based text editing. The method automatically constructs a dataset from an existing table-to-text dataset. 

\subsection{Data Sources}
There are two benchmark datasets of table-to-text, \textsc{WebNLG}~\cite{gardent-etal-2017-creating}\footnote{The data is available at \url{https://github.com/ThiagoCF05/webnlg}. We utilize version 1.5.} and \textsc{RotoWire}\cite{wiseman2017challenges}\footnote{We utilize the \textsc{RotoWire-modified} data provided by~\citet{iso2019learning} available at \url{https://github.com/aistairc/rotowire-modified}. The authors also provide an information extractor for processing the data.}. We create two datasets on the basis of them, referred to as \textsc{WebEdit} and \textsc{RotoEdit} respectively. In the datasets, each instance consists of a table (structured data) and an associated text (unstructured data) describing almost the same content.\footnote{In \textsc{RotoWire}, we discard redundant box-scores and unrelated sentences using the information extractor and heuristic rules. }.  

For each instance, we take the table as triples of facts and the associated text as a revised text, and we automatically create a draft text. The set of triples is represented as $\mathcal{T} = \{t\}$. Each triple $t$ consists of subject, predicate, and object, denoted as $t = (subj, pred, obj)$. For simplicity, we refer to the nouns or noun phrases of subject and object simply as entities. The revised text is a sequence of words denoted as $\boldsymbol{y}$. The draft text is a sequence of words denoted as $\boldsymbol{x}$. 

Given the set of triples $\mathcal{T}$ and the revised text $\boldsymbol{y}$, we aim to create a draft text $\boldsymbol{x}$, such that $\boldsymbol{x}$ is not in accordance with $\mathcal{T}$, in contrast to $\boldsymbol{y}$, and therefore text editing from $\boldsymbol{x}$ to $\boldsymbol{y}$ is needed.

\subsection{Procedure}

Our method first creates templates for all the sets of triples and revised texts and then constructs a draft text for each set of triples and revised text based on their related templates.

\subsubsection*{Creation of templates}

For each instance, our method first delexicalizes the entity words in the set of triples $\mathcal{T}$ and the revised text $\boldsymbol{y}$ to obtain a set of triple templates $\mathcal{T}'$ and a revised template $\boldsymbol{y}'$. For example, given $\mathcal{T} =$\{(Baymax, voice, Scott\_Adsit)\} and $\boldsymbol{y} = $``Scott\_Adsit does the voice for Baymax'', it produces the set of triple templates $\mathcal{T}'=$\{(AGENT-1, voice, PATIENT-1)\} and the revised template $\boldsymbol{y}'=$``AGENT-1 does the voice for PATIENT-1''.
Our method then collects all the sets of triple templates $\mathcal{T}'$ and revised templates $\boldsymbol{y}'$ and retains them in a key-value store with $\boldsymbol{y}'$ being a key and $\mathcal{T}'$ being a value.

\subsubsection*{Creation of draft text}

Next, our method constructs a draft text $\boldsymbol{x}$ using a set of triple templates $\mathcal{T}'$ and a revised template $\boldsymbol{y}'$. For simplicity, it only considers the use of either \textit{insertion} or \textit{deletion} in the text editing, and one can easily make an extension of it to a more complex setting.  Note that the process of data creation is reverse to that of text editing.

Given a pair of $\mathcal{T}'$ and $\boldsymbol{y}'$, our method retrieves another pair denoted as $\hat{\mathcal{T}}'$ and $\hat{x}'$, such that $\boldsymbol{y}'$ and  $\hat{\boldsymbol{x}}'$ have the longest common subsequences. We refer to $\hat{\boldsymbol{x}}'$ as a reference template. There are two possibilities; $\hat{\mathcal{T}}'$ is a subset or a superset of $\mathcal{T}'$. (We ignore the case in which they are identical.) Our method then manages to change $\boldsymbol{y}'$ to a draft template denoted as $\boldsymbol{x}'$ on the basis of the relation between $\mathcal{T}'$ and $\hat{\mathcal{T}}'$. If $\hat{\mathcal{T}}' \subsetneq \mathcal{T}'$, then the draft template $\boldsymbol{x}'$ created is for insertion, and if $\hat{\mathcal{T}}' \supsetneq \mathcal{T}'$, then the draft template $\boldsymbol{x}'$ created is for deletion.

For \textit{insertion}, the revised template $\boldsymbol{y}'$ and the reference template $\hat{\boldsymbol{x}}'$ share subsequences, and the set of triples $\mathcal{T}\backslash\hat{\mathcal{T}}$ appear in $\boldsymbol{y}'$ but not in $\hat{\boldsymbol{x}}'$.  Our method keeps the shared subsequences in $\boldsymbol{y}'$, removes the subsequences in $\boldsymbol{y}'$ about $\mathcal{T}\backslash\hat{\mathcal{T}}$, and copies the rest of words in $\boldsymbol{y}'$, to create the draft template $\boldsymbol{x}'$. Table~\ref{tab:insert} gives an example. The shared subsequences ``AGENT-1 performed as PATIENT-3 on BRIDGE-1 mission'' are kept. The set of triple templates $\mathcal{T}\backslash\hat{\mathcal{T}}$ is \{(BRIDGE-1, operator, PATIENT-2)\}. The subsequence ``that was operated by PATIENT-2'' is removed. Note that the subsequence ``served'' is not copied because it is not shared by $\boldsymbol{y}'$ and $\hat{\boldsymbol{x}}'$.

For \textit{deletion}, the revised template $\boldsymbol{y}'$ and the reference template $\hat{\boldsymbol{x}}'$ share subsequences. The set of triples $\hat{\mathcal{T}}\backslash\mathcal{T}$ appear in $\hat{\boldsymbol{x}}'$ but not in $\boldsymbol{y}'$.  Our method retains the shared subsequences in $\boldsymbol{y}'$, copies the subsequences in $\hat{\boldsymbol{x}}'$ about $\hat{\mathcal{T}}\backslash\mathcal{T}$, and copies the rest of words in $\boldsymbol{y}'$, to create the draft template $\boldsymbol{x}'$. Table~\ref{tab:delete} gives an example. The subsequences ``AGENT-1 was created by BRIDGE-1 and PATIENT-2'' are retained. The set of triple templates $\hat{\mathcal{T}}\backslash\mathcal{T}$ is \{(AGENT-1, fullName, PATIENT-1)\}. The subsequence ``whose full name is PATIENT-1'' is copied. Note that the subsequence ``the character of'' is not copied because it is not shared by $\boldsymbol{y}'$ and $\hat{\boldsymbol{x}}'$.

After getting the draft template $\boldsymbol{x}'$, our method lexicalizes it to obtain a draft text $\boldsymbol{x}$, where the lexicons (entity words) are collected from the corresponding revised text $\boldsymbol{y}$.

We obtain two datasets with our method, referred to as \textsc{WebEdit} and \textsc{RotoEdit}, respectively.
Table~\ref{tab:stats} gives the statistics of the datasets.

In the \textsc{WebEdit} data, sometimes entities only appear in the $subj$'s of triples. In such cases, we also make them appear in the $obj$'s. To do so, we introduce an additional triple (\texttt{ROOT}, \texttt{IsOf}, $subj$) for each $subj$, where \texttt{ROOT} is a dummy entity.

\begin{table}[t]
    \centering
    \footnotesize
    \begin{tabular}{@{}c|cccccc@{}}
        \toprule
         & \multicolumn{3}{c}{\textsc{WebEdit}} & \multicolumn{3}{c}{\textsc{RotoEdit}}\\
         & \textsc{\scriptsize{Train}} & \textsc{\scriptsize{Valid}} & \textsc{\scriptsize{Test}}& \textsc{\scriptsize{Train}} & \textsc{\scriptsize{Valid}} & \textsc{\scriptsize{Test}}\\
          \midrule
        $\#\mathcal{D}$ & 181k & 23k & 29k & 27k & 5.3k & 4.9k\\
        $\#\mathcal{W}_\texttt{d}$ & 4.1M & 495k & 624k & 4.7M & 904k & 839k\\
        $\#\mathcal{W}_\texttt{r}$ & 4.2M & 525k & 649k & 5.6M & 1.1M & 1.0M\\
        $\#\mathcal{S}$ & 403k & 49k & 62k & 209k  & 40k  & 36k \\
        \bottomrule
    \end{tabular}
    \caption{Statistics of \textsc{WebEdit} and \textsc{RotoEdit}, where \#$\mathcal{D}$ is the number of instances, \#$\mathcal{W}_\texttt{d}$ and \#$\mathcal{W}_\texttt{r}$ are the total numbers of words in the draft texts and the revised texts, respectively, and \#$\mathcal{S}$ is total the number of sentences.}
    \label{tab:stats}
\end{table}

\section{\textsc{FactEditor}}
\label{sec:model}
In this section, we describe our proposed model for fact-based text editing referred to as \textsc{FactEditor}.

\subsection{Model Architecture}

\textsc{FactEditor} transforms a draft text into a revised text based on given triples. The model consists of three components, a buffer for storing the draft text and its representations, a stream for storing the revised text and its representations, and a memory for storing the triples and their representations, as shown in Figure~\ref{fig:model}.

\textsc{FactEditor} scans the text in the buffer, copies the parts of text from the buffer into the stream if they are described in the triples in the memory, deletes the parts of the text if they are not mentioned in the triples, and inserts new parts of next into the stream which is only presented in the triples.

The architecture of \textsc{FactEditor} is inspired by those in sentence parsing~\citet{dyer2015transition,watanabe-sumita-2015-transition}. The actual processing of \textsc{FactEditor} is to generate a sequence of words into the stream from the given sequence of words in the buffer and the set of triples in the memory.  A neural network is employed to control the entire editing process.

\subsection{Neural Network}
\label{sec:components}
\subsubsection*{Initialization}

\textsc{FactEditor} first initializes the representations of content in the buffer, stream, and memory.

There is a feed-forward network associated with the memory, utilized to create the embeddings of triples. Let $M$ denote the number of triples. The embedding of triple $t_j, j=1,\cdots,M$ is calculated as 
\begin{align*}
    \boldsymbol{t}_j = \tanh (\boldsymbol{W}_{\texttt{t}} \cdot [\boldsymbol{e}_j^{\texttt{subj}}; \boldsymbol{e}_j^{\texttt{pred}} ; \boldsymbol{e}_j^{\texttt{obj}}] + \boldsymbol{b}_{\texttt{t}}),
\end{align*}
where $\boldsymbol{W}_{\texttt{t}}$ and $\boldsymbol{b}_{\texttt{t}}$ denote parameters, $\boldsymbol{e}_j^{\texttt{subj}}, \boldsymbol{e}_j^{\texttt{pred}}, \boldsymbol{e}_j^{\texttt{obj}}$ denote the embeddings of subject, predicate, and object of triple $t_j$, and $[\ ;]$ denotes vector concatenation.

There is a bi-directional \textsc{LSTM} associated with the buffer, utilized to create the embeddings of words of draft text. The embeddings are obtained as $\boldsymbol{b} = \textsc{BiLSTM}(\boldsymbol{x})$,
where $\boldsymbol{x} = (\boldsymbol{x}_1, \dots, \boldsymbol{x}_{N}) $ is the list of embeddings of words and $\boldsymbol{b} = (\boldsymbol{b}_1, \dots, \boldsymbol{b}_{N})$ is the list of representations of words, where $N$ denotes the number of words.

There is an LSTM associated with the stream for representing the hidden states of the stream. The first 
hidden state is initialized as
\begin{align*}
    \boldsymbol{s}_1 = \tanh\left(\boldsymbol{W}_{\texttt{s}} \cdot \left[\frac{\sum_{i=1}^{N}
    \boldsymbol{b}_i}{N} \mathlarger{;} \frac{\sum_{j = 1}^{M} \boldsymbol{t}_j}{M} \right] + \boldsymbol{b}_{\texttt{s}}\right)
\end{align*}
where $\boldsymbol{W}_{\texttt{s}}$ and $\boldsymbol{b}_{\texttt{s}}$ denotes parameters.

\subsubsection*{Action prediction}

\textsc{FactEditor} predicts an action at each time $t$ using the LSTM. There are three types of action, namely \texttt{Keep}, \texttt{Drop}, and \texttt{Gen}. First, it composes a context vector $\tilde{\boldsymbol{t}}_t$ of triples at time $t$ using attention
\begin{align*}
    \tilde{\boldsymbol{t}}_t &= \sum_{j=1}^M \alpha_{t, j} \boldsymbol{t}_j
\end{align*}
where $\alpha_{t, j}$ is a weight calculated as
\begin{align*}
    \alpha_{t, j} & \propto \exp \left(\boldsymbol{v}_{\alpha}^\top \cdot \tanh{(\boldsymbol{W}_{\alpha} \cdot [\boldsymbol{s}_t;\boldsymbol{b}_t;\boldsymbol{t}_j]
    )}\right)
\end{align*}
where $\boldsymbol{v}_{\alpha}$ and $\boldsymbol{W}_{\alpha}$ are parameters. Then, it creates the hidden state $\boldsymbol{z}_t$ for action prediction at time $t$
\begin{align*}
    \boldsymbol{z}_t = \tanh \left(\boldsymbol{W}_{z} \cdot [\boldsymbol{s}_t; \boldsymbol{b}_t; \tilde{\boldsymbol{t}}_t] + \boldsymbol{b}_{z} \right)
\end{align*}
where $\boldsymbol{W}_{z}$ and $\boldsymbol{b}_{z}$ denote parameters. Next, it calculates the probability of action $a_t$
\begin{align*}
    P({a}_t \mid \boldsymbol{z}_t) = \text{softmax}(\boldsymbol{W}_{a} \cdot \boldsymbol{z}_t)
\end{align*}
where $\boldsymbol{W}_{a}$ denotes parameters, and chooses the action having the largest probability.

\subsubsection*{Action execution}

\textsc{FactEditor} takes action based on the prediction result at time $t$.

\begin{figure}[t]
    \centering
    \begin{subfigure}[b]{0.99\linewidth}
        \centering
        \includegraphics[width=\textwidth]{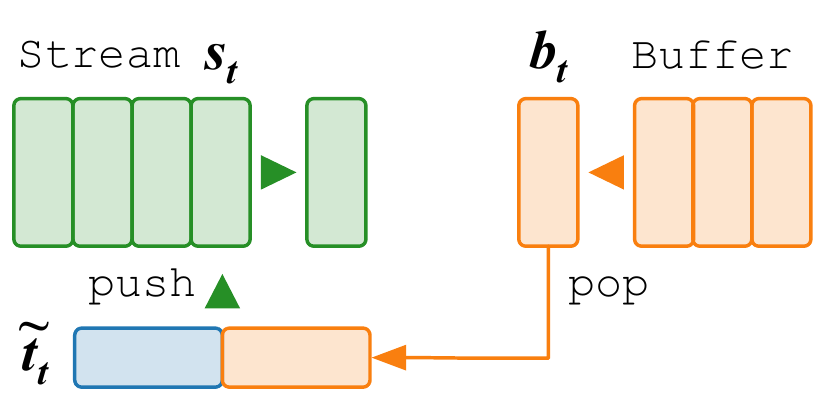}
        \caption{The \texttt{Keep} action, where the top embedding of the buffer $\boldsymbol{b}_t$ is popped and the concatenated vector $[\tilde{\boldsymbol{t}}_t;\boldsymbol{b}_t]$ is pushed into the stream \textsc{LSTM}.}
        \label{fig:keep}
    \end{subfigure}
    \hfill
    \begin{subfigure}[b]{0.99\linewidth}
        \centering
        \includegraphics[width=\textwidth]{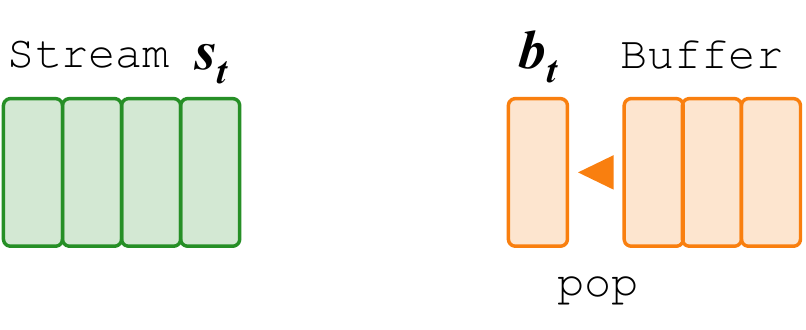}
        \caption{The \texttt{Drop} action, where the top embedding of the buffer $\boldsymbol{b}_t$ is popped and the state in the stream is reused at the next time step $t+1$.}
        \label{fig:drop}
    \end{subfigure}
    \hfill
    \begin{subfigure}[b]{0.99\linewidth}
        \centering
        \includegraphics[width=\textwidth]{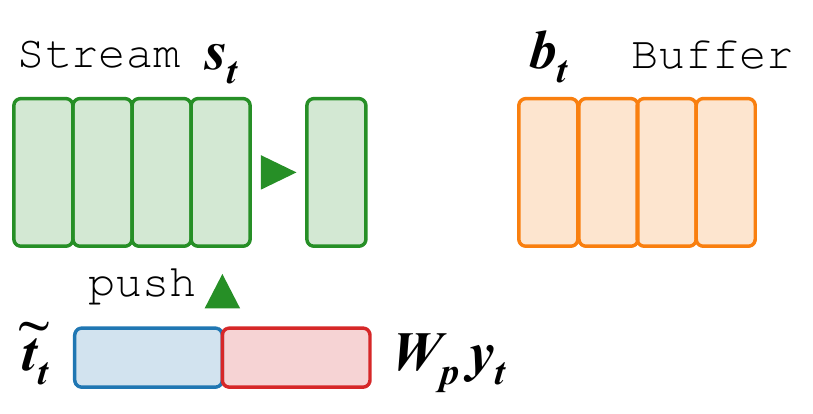}
        \caption{The \texttt{Gen} action, where the concatenated vector $[\tilde{\boldsymbol{t}}_t;\boldsymbol{W}_p\boldsymbol{y}_t]$ is pushed into the stream, and the top embedding of the buffer is reused at the next time step $t+1$.}
        \label{fig:gen}
    \end{subfigure}
    \caption{Actions of \textsc{FactEditor}.}
    \label{fig:model}
\end{figure}

\begin{table*}[t]
    \centering
    \small
    \begin{tabular}{r|l}
        \toprule
        Draft text $\boldsymbol{x}$ & \green{Bakewell\_pudding} \green{is} \green{Dessert} \green{that} \orange{can be served Warm or cold} \green{.} \\\midrule
        Revised text $\boldsymbol{y}$ & \green{Bakewell\_pudding is Dessert that} \blue{originates from Derbyshire\_Dales} \green{.}\\\midrule
        \multirow{2}{*}{Action sequence $\boldsymbol{a}$} & \green{\texttt{Keep} \texttt{Keep} \texttt{Keep} \texttt{Keep}} \blue{\texttt{Gen}(originates) \texttt{Gen}(from) \texttt{Gen}(Derbyshire\_Dales)}\\
        & \orange{\texttt{Drop} \texttt{Drop} \texttt{Drop} \texttt{Drop}} \green{\texttt{Keep}}\\
        \bottomrule
    \end{tabular}
    \caption{An example of action sequence derived from a draft text and revised text.}
    \label{tab:action}
\end{table*}
For \texttt{Keep} at time \textit{t}, \textsc{FactEditor} pops the top embedding $\boldsymbol{b}_t$ in the buffer, and feeds the combination of the top embedding $\boldsymbol{b}_t$ and the context vector of triples $\tilde{\boldsymbol{t}}_t$ into the stream, as shown in Fig.~\ref{fig:keep}. The state of stream is updated with the \textsc{LSTM} as 
$\boldsymbol{s}_{t+1} = \textsc{LSTM}([\tilde{\boldsymbol{t}}_t;\boldsymbol{b}_t], \boldsymbol{s}_{t})$.  \textsc{FactEditor} also copies the top word in the buffer into the stream.

For \texttt{Drop} at time \textit{t}, \textsc{FactEditor} pops the top embedding in the buffer and proceeds to the next state, as shown in Fig.~\ref{fig:drop}. The state of stream is updated as $\boldsymbol{s}_{t+1} = \boldsymbol{s}_{t}$. Note that no word is inputted into the stream.

For \texttt{Gen} at time \textit{t}, \textsc{FactEditor} does not pop the top embedding in the buffer. It feeds the combination of the context vector of triples $\tilde{\boldsymbol{t}}_t$ and the linearly projected embedding of word $w$ into the stream, as shown in Fig.~\ref{fig:gen}. The state of stream is updated with the LSTM as
$\boldsymbol{s}_{t+1} = \textsc{LSTM}([\tilde{\boldsymbol{t}}_t;\boldsymbol{W}_p \boldsymbol{y}_t], \boldsymbol{s}_{t})$, 
where $\boldsymbol{y}_t$ is the embedding of the generated word $y_t$ and $\boldsymbol{W}_p$ denotes parameters. In addition, \textsc{FactEditor} copies the generated word $y_t$ into the stream.

\textsc{FactEditor} continues the actions until the buffer becomes empty.

\subsubsection*{Word generation}

\textsc{FactEditor} generates a word $y_t$ at time $t$, when the action is \texttt{Gen},
\begin{align*}
    P_{\texttt{gen}}({y}_t \mid \boldsymbol{z}_t) = \text{softmax}(\boldsymbol{W}_{y} \cdot \boldsymbol{z}_t)
\end{align*}
where $\boldsymbol{W}_{y}$ is parameters.

To avoid generation of OOV words, \textsc{FactEditor} exploits the copy mechanism. It calculates the probability of copying the object of triple $t_j$
\begin{align*}
    P_{\texttt{copy}}(o_j \mid \boldsymbol{z}_t) &\propto \exp{(\boldsymbol{v}_{c}^\top \cdot \tanh(\boldsymbol{W}_{c} \cdot [\boldsymbol{z}_t;\boldsymbol{t}_j] ))}
\end{align*}
where $\boldsymbol{v}_{c}$ and $\boldsymbol{W}_{c}$ denote parameters, and $o_j$ is the object of triple $t_j$. It also calculates the probability of gating
\begin{align*}
    p_{\texttt{gate}} = \text{sigmoid}(\boldsymbol{w}_{\texttt{g}}^\top \cdot \boldsymbol{z}_t + b_{\texttt{g}})
\end{align*}
where $\boldsymbol{w}_{\texttt{g}}$ and $b_\texttt{g}$ are parameters. Finally, it calculates the probability of generating a word $w_t$ through either generation or copying,
\begin{multline*}
    P(y_t \mid \boldsymbol{z}_t) = p_{\texttt{gate}} P_{\texttt{gen}}(y_t \mid \boldsymbol{z}_t) \\+ (1 - p_{\texttt{gate}}) \sum_{j=1: o_j = y_t}^{M} P_{\texttt{copy}}(o_j \mid \boldsymbol{z}_t),
\end{multline*}
where it is assumed that the triples in the memory have the same subject and thus only objects need to be copied.

\subsection{Model Learning}

The conditional probability of sequence of actions $\boldsymbol{a} = (a_1, a_2, \cdots, a_T)$ given the set of triples $\mathcal{T}$ and the sequence of input words $\boldsymbol{x}$ can be written as
\begin{align*}
    P(\boldsymbol{a} \mid \mathcal{T}, \boldsymbol{x}) = \prod_{t=1}^{T} P(a_t \mid \boldsymbol{z}_t)
\end{align*}
where $P(a_t \mid \boldsymbol{z}_t)$ is the conditional probability of action $a_t$ given state $\boldsymbol{z}_t$ at time $t$ and $T$ denotes the number of actions.

The conditional probability of sequence of generated words $\boldsymbol{y} = (y_1, y_2, \cdots, y_{T})$ given the sequence of actions $\boldsymbol{a}$ can be written as
\begin{align*}
    P(\boldsymbol{y} \mid \boldsymbol{a}) = \prod_{t=1}^{T} P(y_t \mid a_t)
\end{align*}
where $P(y_t \mid a_t)$ is the conditional probability of generated word $y_t$ given action $a_t$ at time $t$, which
is calculated as
\begin{align*}
  &P(y_t \mid a_t) =
    &\begin{cases}
      P(y_t \mid \boldsymbol{z_t}) & \text{if $a_t =\ $\texttt{Gen}} \\
     1 & \text{otherwise}
    \end{cases}       
\end{align*}
Note that not all positions have a generated word. In such a case, $y_t$ is simply a null word.

The learning of the model is carried out via supervised learning. 
The objective of learning is to minimize the negative log-likelihood of $P(\boldsymbol{a}  \mid \mathcal{T}, \boldsymbol{x})$ and $P(\boldsymbol{y} \mid \boldsymbol{a})$
\begin{align*}
    \mathcal{L}(\boldsymbol{\theta}) = - \sum_{t=1}^{T} \left\{ \log P(a_t \mid \boldsymbol{z}_t) + \log P(y_t \mid a_t)\right\}
\end{align*}
where $\boldsymbol{\theta}$ denotes the parameters.

A training instance consists of a pair of draft text and revised text, as well as a set of triples, denoted as $\boldsymbol{x}$, $\boldsymbol{y}$, and $\mathcal{T}$  respectively. For each instance, our method derives a sequence of actions denoted as $\boldsymbol{a}$, in a similar way as that in~\cite{dong2019editnts}. It first finds the longest common sub-sequence between $\boldsymbol{x}$ and $\boldsymbol{y}$, and then selects an action of \texttt{Keep}, \texttt{Drop}, or \texttt{Gen} at each position, according to how $\boldsymbol{y}$ is obtained from $\boldsymbol{x}$ and $\mathcal{T}$ (cf., Tab.~\ref{tab:action}). Action \texttt{Gen} is preferred over action \texttt{Drop} when both are valid.
\begin{figure*}
    \centering
    \begin{subfigure}[b]{0.3\linewidth}
        \centering
        \includegraphics[height=2.2cm]{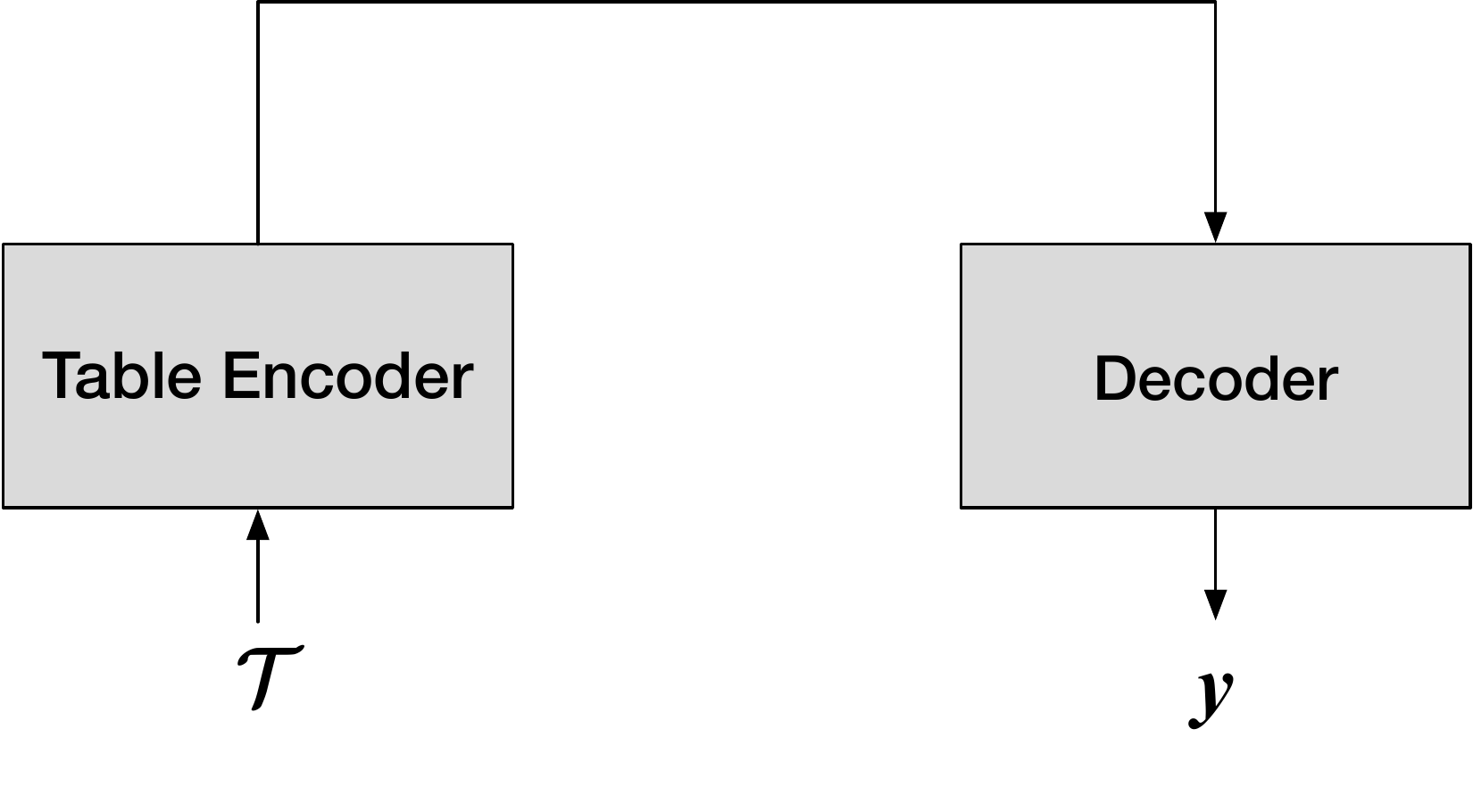}
        \caption{Table-to-Text}
        \label{fig:d2t}
    \end{subfigure}
    \hfill
    \begin{subfigure}[b]{0.3\linewidth}
        \centering
        \includegraphics[height=2.2cm]{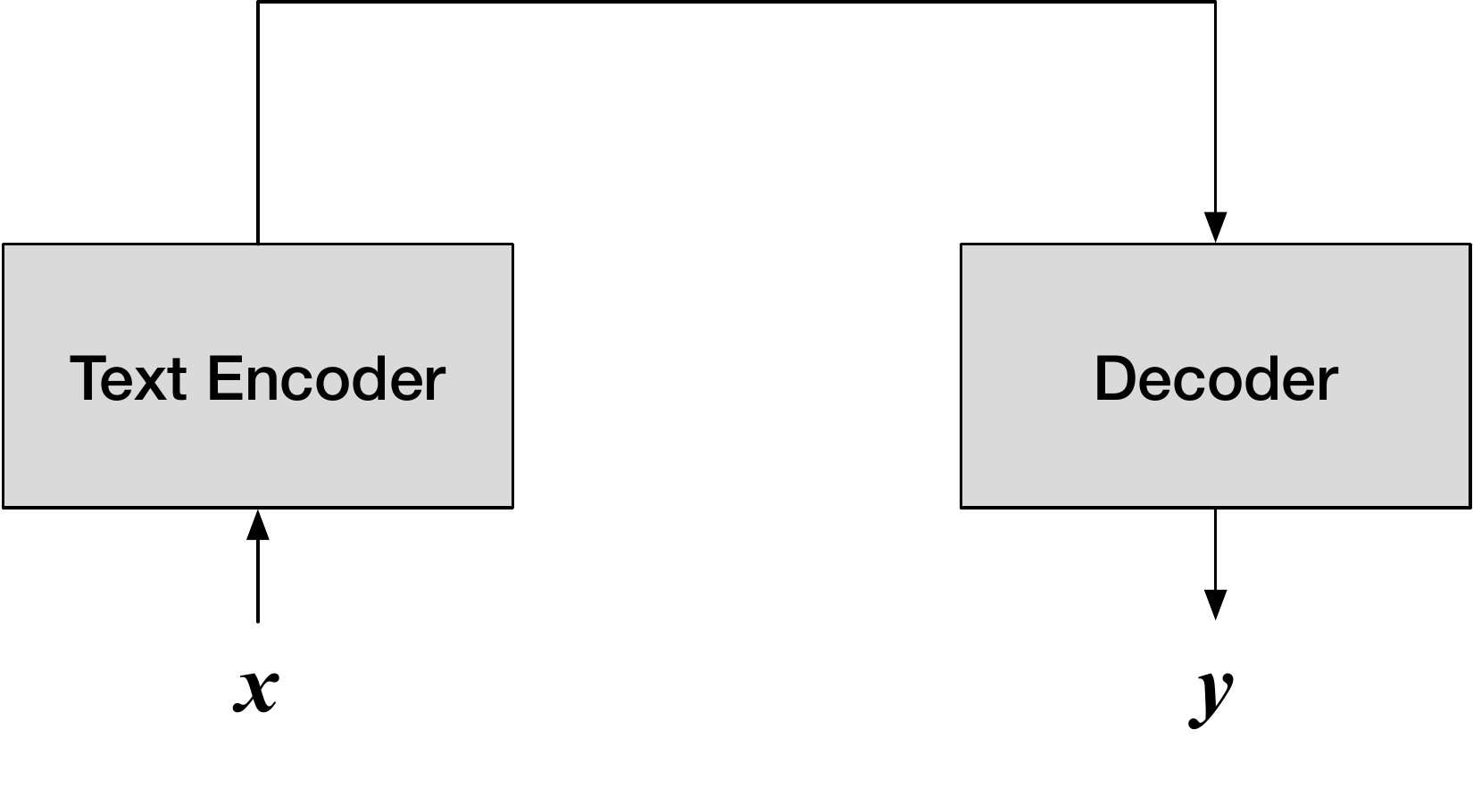}
        \caption{Text-to-Text}
        \label{fig:t2t}
    \end{subfigure}
    \hfill
    \begin{subfigure}[b]{0.35\linewidth}
        \centering
        \includegraphics[height=2.2cm]{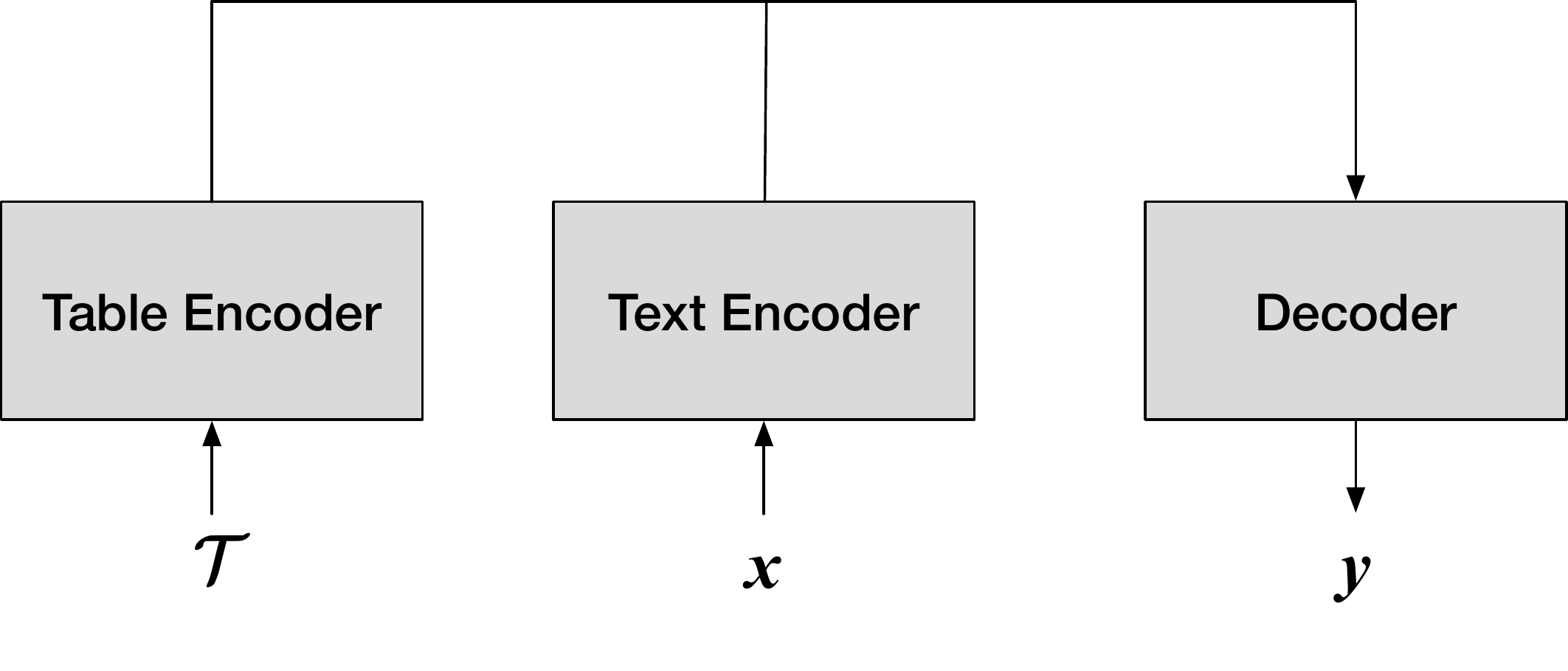}
        \caption{\textsc{EncDecEditor}}
        \label{fig:encdec}
    \end{subfigure}
    \caption{Model architectures of the baselines. All models employ attention and copy mechanism.}
    \label{fig:baselines}
\end{figure*}

\subsection{Time Complexity}
\label{sec:runtime}

The time complexity of inference in \textsc{FactEditor} is $\mathcal{O}(NM)$, where $N$ is the number of words in the buffer, and $M$ is the number of triples. Scanning of data in the buffer is of complexity $\mathcal{O}(N)$. The generation of action needs the execution of attention, which is of complexity $\mathcal{O}(M)$. Usually, $N$ is much larger than $M$. 

\subsection{Baseline}

We consider a baseline method using the encoder-decoder architecture, which takes the set of triples and the draft text as input and generates a revised text. We refer to the method as \textsc{EncDecEditor}. The encoder of \textsc{EncDecEditor} is the same as that of \textsc{FactEditor}.  The decoder is the standard attention and copy model, which creates and utilizes a context vector and predicts the next word at each time.

The time complexity of inference in \textsc{EncDecEditor} is $\mathcal{O}(N^2 + NM)$ (cf.,\citet{britz2017efficient}). Note that in fact-based text editing, usually $N$ is very large. That means that \textsc{EncDecEditor} is less efficient than \textsc{FactEditor}.

\begin{table*}[t]
    \begin{subtable}[t]{\textwidth}
        \centering
        \small
        \begin{tabular}{l|c|cccc|c|ccc}
            \toprule
            \multirow{2}{*}{\textbf{Model}} &  \multicolumn{6}{c|}{\textsc{Fluency}} &  \multicolumn{3}{c}{\textsc{Fidelity}} \\
            &\textsc{Bleu} & \textsc{Sari} & \textsc{Keep} & \textsc{Add} & \textsc{Delete} & \textsc{EM} & P\% & R\% & F1\%\\
            \midrule
            \multicolumn{10}{l}{\textbf{Baselines}}\\
            No-Editing              & 66.67 & 31.51 & 78.62 & 3.91 & 12.02. & 0. & 84.49 & 76.34 & 80.21\\
            Table-to-Text       & 33.75 & 43.83 & 51.44 & 27.86 & 52.19 & 5.78 & \textbf{98.23} & 83.72 & 90.40\\
            Text-to-Text       & 63.61 & 58.73 & 82.62 & 25.77 & 67.80 & 6.22 & 81.93 & 77.16 & 79.48\\
            \midrule
            \multicolumn{10}{l}{\textbf{Fact-based text editing}}\\
            \textsc{EncDecEditor} & 71.03 & 69.59 & 89.49 & 43.82 & 75.48 & 20.96 & 98.06 & 87.56 & 92.51\\
            \textsc{FactEditor} & \textbf{75.68} & \textbf{72.20} & \textbf{91.84} & \textbf{47.69} & \textbf{77.07} & \textbf{24.80} & 96.88 & \textbf{89.74} & \textbf{93.17}\\
            \bottomrule
        \end{tabular}
        \caption{\textsc{WebEdit}}
        \label{tab:Experiment_webedit}
    \end{subtable}
    \newline
    \vspace*{1em}
    \newline
    \begin{subtable}[t]{\textwidth}
        \centering
        \small
        \begin{tabular}{l|c|cccc|c|ccc}
            \toprule
            \multirow{2}{*}{\textbf{Model}} &  \multicolumn{6}{c|}{\textsc{Fluency}} &  \multicolumn{3}{c}{\textsc{Fidelity}} \\
            &\textsc{Bleu} & \textsc{Sari} & \textsc{Keep} & \textsc{Add} & \textsc{Delete} & \textsc{EM} & P\% & R\% & F1\%\\
            \midrule
            \multicolumn{10}{l}{\textbf{Baselines}}\\
            No-Editing              & 74.95 & 39.59 & 95.72 & 0.05 & 23.01 & 0. & 92.92 & 65.02 & 76.51\\
            Table-to-Text      & 24.87 & 23.30 & 39.12 & 14.78 & 16.00 & 0. & 48.01 & 24.28 & 32.33 \\
            Text-to-Text       & 78.07 & 60.25 & 97.29 & 13.04 & 70.43 & 0.02 & 63.62 & 41.08 & 49.92\\
            \midrule
            \multicolumn{10}{l}{\textbf{Fact-based text editing}}\\
            \textsc{EncDecEditor} & 83.36 & 71.46 & 97.69 & \textbf{44.02} & 72.69 & 2.49 & 78.80 & 52.21 & 62.81\\
            \textsc{FactEditor} & \textbf{84.43} & \textbf{74.72} & \textbf{98.41} & 41.50 & \textbf{84.24} & \textbf{2.65} & \textbf{78.84} & \textbf{52.30} & \textbf{63.39}\\
            \bottomrule
        \end{tabular}
        \caption{\textsc{RotoEdit}}
        \label{tab:Experiment_rotoedit}
    \end{subtable}
    \caption{Performances of \textsc{FactEditor} and baselines on two datasets in terms of Fluency and Fidelity. \textsc{EM} stands for exact match.}
    \label{tab:main_results}
\end{table*}

\section{Experiment}

We conduct experiments to make comparison between \textsc{FactEditor} and the baselines using the two datasets \textsc{WebEdit} and \textsc{RotoEdit}.

\subsection{Experiment Setup}

The main baseline is the encoder-decoder model \textsc{EncDecEditor}, as explained above. We further consider three baselines, No-Editing, Table-to-Text, and Text-to-Text. In No-Editing, the draft text is directly used.
In Table-to-Text, a revised text is generated from the triples using encoder-decoder. In Text-to-Text, a revised text is created from the draft text using the encoder-decoder model.
Figure \ref{fig:baselines} gives illustrations of the baselines.

We evaluate the results of revised texts by the models from the viewpoint of fluency and fidelity. We utilize ExactMatch (EM), \textsc{Bleu}~\cite{papineni-etal-2002-bleu} and \textsc{Sari}~\cite{xu2016optimizing} scores\footnote{We use a modified version of \textsc{SARI} where $\beta$ equals $1.0$, available at \url{https://github.com/tensorflow/tensor2tensor/blob/master/tensor2tensor/utils/sari_hook.py}} as evaluation metrics for fluency. We also utilize precision, recall, and F1 score as evaluation metrics for fidelity. For \textsc{WebEdit}, we extract the entities from the generated text and the reference text and then calculate the precision, recall, and F1 scores. For \textsc{RotoEdit}, we use the information extraction tool provided by \citet{wiseman2017challenges} for calculation of the scores.

For the embeddings of subject and object for both datasets and the embedding of the predicate for \textsc{RotoEdit}, we simply use the embedding lookup table. For the embedding of the predicate for \textsc{WebEdit}, we first tokenize the predicate, lookup the embeddings of lower-cased words from the table, and use averaged embedding to deal with the OOV problem~\cite{moryossef2019step}.

We tune the hyperparameters based on the \textsc{Bleu} score on a development set. For \textsc{WebEdit}, we set the sizes of embeddings, buffers, and triples to 300, and set the size of the stream to 600. For \textsc{RotoEdit}, we set the size of embeddings to 100 and set the sizes of buffers, triples, and stream to 200. The initial learning rate is 2e-3, and AMSGrad is used for automatically adjusting the learning rate~\cite{reddi2018convergence}. Our implementation makes use of AllenNLP~\cite{gardner2018allennlp}.

\subsection{Experimental Results}
\label{sec:results}

\subsubsection*{Quantitative evaluation}
We present the performances of our proposed model \textsc{FactEditor} and the baselines on fact-based text editing in Table \ref{tab:main_results}. One can draw several conclusions from the results.

First, our proposed model, \textsc{FactEditor}, achieves significantly better performances than the main baseline, \textsc{EncDecEditor}, in terms of almost all measures. In particular, \textsc{FactEditor} obtains significant gains in \textsc{Delete} scores on both \textsc{WebEdit} and \textsc{RotoEdit}.

Second, the fact-based text editing models (\textsc{FactEditor} and \textsc{EncDecEditor}) significantly improve upon the other models in terms of fluency scores, and achieve similar performances in terms of fidelity scores.

Third, compared to No-Editing, Table-to-Text has higher fidelity scores, but lower fluency scores. Text-to-Text has almost the same fluency scores, but lower fidelity scores on \textsc{RotoEdit}.

\subsubsection*{Qualitative evaluation}
We also manually evaluate 50 randomly sampled revised texts for \textsc{WebEdit}. We check whether the revised texts given by \textsc{FactEditor} and \textsc{EncDecEditor} include all the facts. We categorize the factual errors made by the two models. Table~\ref{tab:error} shows the results. One can see that \textsc{FactEditor} covers more facts than \textsc{EncDecEditor} and has less factual errors than \textsc{EncDecEditor}. 

\begin{table}[t]
    \centering
    \scriptsize
    \setlength\tabcolsep{2.5pt}
    \begin{tabular}{l|cc|cccc}
        \toprule
        & \multicolumn{2}{c|}{\textbf{Covered facts}} & \multicolumn{4}{c}{\textbf{Factual errors}}\\
        & ~\textsc{CQT}  & ~\textsc{UPara} & ~\textsc{Rpt} & ~\textsc{Ms} & ~\textsc{USup} & ~\textsc{DRel} \\\midrule
        \textsc{EncDecEditor}& 14 & 7 & 16 & 21 & 3 & 12\\
        \textsc{FactEditor} & \textbf{24} & \textbf{4} & \textbf{9} &\textbf{19} &\textbf{1} &\textbf{3}\\
        \bottomrule
    \end{tabular}
    \caption{Evaluation results on 50 randomly sampled revised texts in \textsc{WebEdit} in terms of numbers of correct editing (\textsc{CQT}), unnecessary paraphrasing (\textsc{UPara}),  repetition (\textsc{Rpt}), missing facts (\textsc{Ms}), unsupported facts (\textsc{USup}) and different relations (\textsc{DRel})}
    \label{tab:error}
\end{table}

\textsc{FactEditor} has a larger number of correct editing (\textsc{CQT}) than \textsc{EncDecEditor} for fact-based text editing. In contrast, \textsc{EncDecEditor} often includes a larger number of unnecessary rephrasings (\textsc{UPara}) than \textsc{FactEditor}.

There are four types of factual errors: fact repetitions (\textsc{Rpt}), fact missings (\textsc{Ms}), fact unsupported (\textsc{USup}), and relation difference (\textsc{DRel}). Both \textsc{FactEditor} and \textsc{EncDecEditor} often fail to insert missing facts (\textsc{Ms}), but rarely insert unsupported facts (\textsc{USup}). \textsc{EncDecEditor} often generates the same facts multiple times (\textsc{RPT}) or facts in different relations (\textsc{DRel}). In contrast, \textsc{FactEditor} can seldomly make such errors.

\begin{table*}[t]
    \centering
    \scriptsize
    \begin{tabular}{l|l}
    \toprule
        Set of triples & \begin{tabular}{@{}c@{}lll@{}}
        \{&(\green{Ardmore\_Airport}, & \green{runwayLength}, & \green{1411.0}), \\
        &(\green{Ardmore\_Airport}, & \green{3rd\_runway\_SurfaceType}, & \green{Poaceae}), \\
        &(\blue{Ardmore\_Airport}, & \blue{operatingOrganisation}, & \blue{Civil\_Aviation\_Authority\_of\_New\_Zealand}), \\
        &(\green{Ardmore\_Airport}, & \green{elevationAboveTheSeaLevel}, & \green{34.0}), \\
        &(\green{Ardmore\_Airport}, & \green{runwayName}, & \green{03R/21L})\}
        \end{tabular}\\\midrule
        Draft text & 
        \begin{tabular}[c]{@{}l@{}} \green{Ardmore\_Airport} , \orange{ICAO Location Identifier UTAA} . \green{Ardmore\_Airport} 3rd runway\\ is made of \green{Poaceae} and \green{Ardmore\_Airport} . \green{03R/21L} is \green{1411.0} m long and \green{Ardmore\_Airport} \\is \green{34.0} above sea level . \end{tabular}
        \\\midrule
        Revised text & \begin{tabular}[c]{@{}l@{}} \green{Ardmore\_Airport} is operated by \blue{Civil\_Aviation\_Authority\_of\_New\_Zealand} . \green{Ardmore\_Airport}\\ 3rd runway is made of \green{Poaceae} and \green{Ardmore\_Airport} name is \green{03R/21L} . \green{03R/21L} is \green{1411.0} m long\\ and \green{Ardmore\_Airport} is \green{34.0} above sea level .\end{tabular}
\\\midrule
        \textsc{EncDecEditor} & \begin{tabular}[c]{@{}l@{}} \green{Ardmore\_Airport} , \orange{ICAO Location Identifier UTAA} , is operated by \\\blue{Civil\_Aviation\_Authority\_of\_New\_Zealand} . \green{Ardmore\_Airport} 3rd runway is made of \green{Poaceae} and \\\green{Ardmore\_Airport} . \green{03R/21L} is \green{1411.0} m long and \green{Ardmore\_Airport} is \underline{\green{34.0} m long} . \end{tabular}\\\midrule
        \textsc{FactEditor} & \begin{tabular}[c]{@{}l@{}} \green{Ardmore\_Airport} is operated by \blue{Civil\_Aviation\_Authority\_of\_New\_Zealand} . \green{Ardmore\_Airport} \\3rd runway is made of \green{Poaceae} and \green{Ardmore\_Airport} . \green{03R/21L} is \green{1411.0} m long and \\\green{Ardmore\_Airport} is \green{34.0} above sea level .\end{tabular}\\
    \bottomrule
    \end{tabular}
    \caption{Example of generated revised texts given by \textsc{EncDecEditor} and \textsc{FactEditor} on \textsc{WebEdit}. \green{Entities in green} appear in both the set of triples and the draft text. \orange{Entities in orange} only appear in the draft text. \blue{Entities in blue} should appear in the revised text but do not appear in the draft text.}
    \label{tab:example_outputs}
\end{table*}
Table~\ref{tab:example_outputs} shows an example of results
given by \textsc{EncDecEditor} and \textsc{FactEditor}. The revised texts of both \textsc{EncDecEditor} and \textsc{FactEditor} appear to be fluent, but that of \textsc{FactEditor} has higher fidelity than that of \textsc{EncDecEditor}. \textsc{EncDecEditor} cannot effectively eliminate the description about an unsupported fact (in orange) appearing in the draft text. In contrast, \textsc{FactEditor} can deal with the problem well. In addition, \textsc{EncDecEditor} conducts an unnecessary substitution in the draft text (underlined). \textsc{FactEditor} tends to avoid such unnecessary editing.

\subsubsection*{Runtime analysis}
We conduct runtime analysis on \textsc{FactEditor} and the baselines in terms of number of processed words per second, on both \textsc{WebEdit} and \textsc{RotoEdit}. Table~\ref{tab:runtime} gives the results when the batch size is 128 for all methods. Table-to-Text is the fastest, followed by \textsc{FactEditor}. \textsc{FactEditor} is always faster than \textsc{EncDecEditor}, apparently because it has a lower time complexity, as explained in Section 4. The texts in \textsc{WebEdit} are relatively short, and thus \textsc{FactEditor} and \textsc{EncDecEditor} have similar runtime speeds. In contrast, the texts in \textsc{RotoEdit} are relatively long, and thus \textsc{FactEditor} executes approximately two times faster than \textsc{EncDecEditor}.

\begin{table}[t]
    \centering
    \small
    \begin{tabular}{c|cc}
        \toprule
        & \textsc{WebEdit} & \textsc{RotoEdit}\\\midrule
        Table-to-Text & \textbf{4,083} & \textbf{1,834}\\
        Text-to-Text & 2,751 & 581 \\
        \textsc{EncDecEditor} & 2,487 & 505\\
        \textsc{FactEditor} & \underline{3,295} & \underline{1,412}\\
        \bottomrule
    \end{tabular}
    \caption{Runtime analysis (\# of words/second). Table-to-Text always shows the fastest performance (\textbf{Bold-faced}). \textsc{FactEditor} shows the second fastest runtime performance  (\underline{Underlined}).}
    \label{tab:runtime}
\end{table}

\section{Conclusion}
In this paper, we have defined a new task referred to as \textit{fact-based text editing} and made two contributions to research on the problem. First, we have proposed a data construction method for fact-based text editing and created two datasets. Second, we have proposed a model for fact-based text editing, named \textsc{FactEditor}, which performs the task by generating a sequence of actions. Experimental results show that the proposed model \textsc{FactEditor} performs better and faster than the baselines, including an encoder-decoder model.

\section*{Acknowledgments}
We would like to thank the reviewers for their insightful comments. 
\bibliography{acl2020}

\begin{thebibliography}{42}
\expandafter\ifx\csname natexlab\endcsname\relax\def\natexlab#1{#1}\fi

\bibitem[{Bahdanau et~al.(2015)Bahdanau, Cho, and Bengio}]{bahdanau2015neural}
Dzmitry Bahdanau, Kyunghyun Cho, and Yoshua Bengio. 2015.
\newblock \href {https://arxiv.org/abs/1409.0473} {{Neural Machine Translation
  by Jointly Learning to Align and Translate}}.
\newblock In \emph{International Conference on Learning Representations}.

\bibitem[{Britz et~al.(2017)Britz, Guan, and Luong}]{britz2017efficient}
Denny Britz, Melody Guan, and Minh-Thang Luong. 2017.
\newblock \href {https://doi.org/10.18653/v1/D17-1040} {{Efficient Attention
  using a Fixed-Size Memory Representation}}.
\newblock In \emph{Proceedings of the 2017 Conference on Empirical Methods in
  Natural Language Processing}, pages 392--400, Copenhagen, Denmark.
  Association for Computational Linguistics.

\bibitem[{Cho et~al.(2014)Cho, van Merri{\"e}nboer, Gulcehre, Bahdanau,
  Bougares, Schwenk, and Bengio}]{cho2014learning}
Kyunghyun Cho, Bart van Merri{\"e}nboer, Caglar Gulcehre, Dzmitry Bahdanau,
  Fethi Bougares, Holger Schwenk, and Yoshua Bengio. 2014.
\newblock \href {https://doi.org/10.3115/v1/D14-1179} {{Learning Phrase
  Representations using {RNN} Encoder{--}Decoder for Statistical Machine
  Translation}}.
\newblock In \emph{Proceedings of the 2014 Conference on Empirical Methods in
  Natural Language Processing ({EMNLP})}, pages 1724--1734, Doha, Qatar.
  Association for Computational Linguistics.

\bibitem[{Dolan and Brockett(2005)}]{dolan-brockett-2005-automatically}
William~B. Dolan and Chris Brockett. 2005.
\newblock \href {https://www.aclweb.org/anthology/I05-5002} {Automatically
  constructing a corpus of sentential paraphrases}.
\newblock In \emph{Proceedings of the Third International Workshop on
  Paraphrasing ({IWP}2005)}.

\bibitem[{Dong et~al.(2019)Dong, Li, Rezagholizadeh, and
  Cheung}]{dong2019editnts}
Yue Dong, Zichao Li, Mehdi Rezagholizadeh, and Jackie Chi~Kit Cheung. 2019.
\newblock \href {https://doi.org/10.18653/v1/P19-1331} {{{E}dit{NTS}: An Neural
  Programmer-Interpreter Model for Sentence Simplification through Explicit
  Editing}}.
\newblock In \emph{Proceedings of the 57th Annual Meeting of the Association
  for Computational Linguistics}, pages 3393--3402, Florence, Italy.
  Association for Computational Linguistics.

\bibitem[{Dyer et~al.(2015)Dyer, Ballesteros, Ling, Matthews, and
  Smith}]{dyer2015transition}
Chris Dyer, Miguel Ballesteros, Wang Ling, Austin Matthews, and Noah~A. Smith.
  2015.
\newblock \href {https://doi.org/10.3115/v1/P15-1033} {{Transition-Based
  Dependency Parsing with Stack Long Short-Term Memory}}.
\newblock In \emph{Proceedings of the 53rd Annual Meeting of the Association
  for Computational Linguistics and the 7th International Joint Conference on
  Natural Language Processing (Volume 1: Long Papers)}, pages 334--343,
  Beijing, China. Association for Computational Linguistics.

\bibitem[{Gardent et~al.(2017)Gardent, Shimorina, Narayan, and
  Perez-Beltrachini}]{gardent-etal-2017-creating}
Claire Gardent, Anastasia Shimorina, Shashi Narayan, and Laura
  Perez-Beltrachini. 2017.
\newblock \href {https://doi.org/10.18653/v1/P17-1017} {Creating training
  corpora for {NLG} micro-planners}.
\newblock In \emph{Proceedings of the 55th Annual Meeting of the Association
  for Computational Linguistics (Volume 1: Long Papers)}, pages 179--188,
  Vancouver, Canada. Association for Computational Linguistics.

\bibitem[{Gardner et~al.(2018)Gardner, Grus, Neumann, Tafjord, Dasigi, Liu,
  Peters, Schmitz, and Zettlemoyer}]{gardner2018allennlp}
Matt Gardner, Joel Grus, Mark Neumann, Oyvind Tafjord, Pradeep Dasigi,
  Nelson~F. Liu, Matthew Peters, Michael Schmitz, and Luke Zettlemoyer. 2018.
\newblock \href {https://doi.org/10.18653/v1/W18-2501} {{{A}llen{NLP}: A Deep
  Semantic Natural Language Processing Platform}}.
\newblock In \emph{Proceedings of Workshop for {NLP} Open Source Software
  ({NLP}-{OSS})}, pages 1--6, Melbourne, Australia. Association for
  Computational Linguistics.

\bibitem[{Gatt and Krahmer(2018)}]{gatt2018survey}
Albert Gatt and Emiel Krahmer. 2018.
\newblock \href {http://arxiv.org/abs/https://arxiv.org/abs/1703.09902}
  {{Survey of the State of the Art in Natural Language Generation: Core tasks,
  applications and evaluation}}.
\newblock \emph{Journal of Artificial Intelligence Research (JAIR)},
  61:65--170.

\bibitem[{Gu et~al.(2016)Gu, Lu, Li, and Li}]{gu-etal-2016-incorporating}
Jiatao Gu, Zhengdong Lu, Hang Li, and Victor~O.K. Li. 2016.
\newblock \href {https://doi.org/10.18653/v1/P16-1154} {Incorporating copying
  mechanism in sequence-to-sequence learning}.
\newblock In \emph{Proceedings of the 54th Annual Meeting of the Association
  for Computational Linguistics (Volume 1: Long Papers)}, pages 1631--1640,
  Berlin, Germany. Association for Computational Linguistics.

\bibitem[{Gulcehre et~al.(2016)Gulcehre, Ahn, Nallapati, Zhou, and
  Bengio}]{gulcehre-etal-2016-pointing}
Caglar Gulcehre, Sungjin Ahn, Ramesh Nallapati, Bowen Zhou, and Yoshua Bengio.
  2016.
\newblock \href {https://doi.org/10.18653/v1/P16-1014} {Pointing the unknown
  words}.
\newblock In \emph{Proceedings of the 54th Annual Meeting of the Association
  for Computational Linguistics (Volume 1: Long Papers)}, pages 140--149,
  Berlin, Germany. Association for Computational Linguistics.

\bibitem[{Guu et~al.(2018)Guu, Hashimoto, Oren, and Liang}]{guu2018generating}
Kelvin Guu, Tatsunori~B Hashimoto, Yonatan Oren, and Percy Liang. 2018.
\newblock \href {https://www.aclweb.org/anthology/Q18-1031} {{Generating
  Sentences by Editing Prototypes}}.
\newblock \emph{Transactions of the Association for Computational Linguistics},
  6:437--450.

\bibitem[{Hashimoto et~al.(2018)Hashimoto, Guu, Oren, and
  Liang}]{hashimoto2018retrieve}
Tatsunori~B Hashimoto, Kelvin Guu, Yonatan Oren, and Percy~S Liang. 2018.
\newblock \href
  {http://papers.nips.cc/paper/8209-a-retrieve-and-edit-framework-for-predicting-structured-outputs.pdf}
  {{A Retrieve-and-Edit Framework for Predicting Structured Outputs}}.
\newblock In \emph{Advances in Neural Information Processing Systems}, pages
  10052--10062. Curran Associates, Inc.

\bibitem[{Hu et~al.(2017)Hu, Yang, Liang, Salakhutdinov, and
  Xing}]{hu2017toward}
Zhiting Hu, Zichao Yang, Xiaodan Liang, Ruslan Salakhutdinov, and Eric~P. Xing.
  2017.
\newblock \href {http://proceedings.mlr.press/v70/hu17e.html} {{Toward
  Controlled Generation of Text}}.
\newblock In \emph{Proceedings of the 34th International Conference on Machine
  Learning}, volume~70 of \emph{Proceedings of Machine Learning Research},
  pages 1587--1596, International Convention Centre, Sydney, Australia. PMLR.

\bibitem[{Inui et~al.(2003)Inui, Fujita, Takahashi, Iida, and
  Iwakura}]{inui-etal-2003-text}
Kentaro Inui, Atsushi Fujita, Tetsuro Takahashi, Ryu Iida, and Tomoya Iwakura.
  2003.
\newblock \href {https://doi.org/10.3115/1118984.1118986} {Text simplification
  for reading assistance: A project note}.
\newblock In \emph{Proceedings of the Second International Workshop on
  Paraphrasing}, pages 9--16, Sapporo, Japan. Association for Computational
  Linguistics.

\bibitem[{Iso et~al.(2019)Iso, Uehara, Ishigaki, Noji, Aramaki, Kobayashi,
  Miyao, Okazaki, and Takamura}]{iso2019learning}
Hayate Iso, Yui Uehara, Tatsuya Ishigaki, Hiroshi Noji, Eiji Aramaki, Ichiro
  Kobayashi, Yusuke Miyao, Naoaki Okazaki, and Hiroya Takamura. 2019.
\newblock \href {https://www.aclweb.org/anthology/P19-1202} {{Learning to
  Select, Track, and Generate for Data-to-Text}}.
\newblock In \emph{Proceedings of the Annual Meeting of the Association for
  Computational Linguistics (ACL)}, pages 2102--2113, Florence, Italy.

\bibitem[{Knight and Chander(1994)}]{knight1994automated}
Kevin Knight and Ishwar Chander. 1994.
\newblock \href {https://www.aaai.org/Papers/AAAI/1994/AAAI94-119.pdf}
  {{Automated Postediting of Documents}}.
\newblock In \emph{Proceedings of the AAAI Conference on Artificial
  Intelligence.}, volume~94, pages 779--784.

\bibitem[{Lebret et~al.(2016)Lebret, Grangier, and Auli}]{lebret2016neural}
R{\'e}mi Lebret, David Grangier, and Michael Auli. 2016.
\newblock \href {https://doi.org/10.18653/v1/D16-1128} {{Neural Text Generation
  from Structured Data with Application to the Biography Domain}}.
\newblock In \emph{Proceedings of the 2016 Conference on Empirical Methods in
  Natural Language Processing}, pages 1203--1213, Austin, Texas. Association
  for Computational Linguistics.

\bibitem[{Li et~al.(2018)Li, Jiang, Shang, and Li}]{li2018paraphrase}
Zichao Li, Xin Jiang, Lifeng Shang, and Hang Li. 2018.
\newblock \href {https://doi.org/10.18653/v1/D18-1421} {{Paraphrase Generation
  with Deep Reinforcement Learning}}.
\newblock In \emph{Proceedings of the 2018 Conference on Empirical Methods in
  Natural Language Processing}, pages 3865--3878, Brussels, Belgium.
  Association for Computational Linguistics.

\bibitem[{Malmi et~al.(2019)Malmi, Krause, Rothe, Mirylenka, and
  Severyn}]{malmi2019lasertagger}
Eric Malmi, Sebastian Krause, Sascha Rothe, Daniil Mirylenka, and Aliaksei
  Severyn. 2019.
\newblock \href {https://doi.org/10.18653/v1/D19-1510} {{Encode, Tag, Realize:
  High-Precision Text Editing}}.
\newblock In \emph{Proceedings of the 2019 Conference on Empirical Methods in
  Natural Language Processing and the 9th International Joint Conference on
  Natural Language Processing (EMNLP-IJCNLP)}, pages 5053--5064, Hong Kong,
  China. Association for Computational Linguistics.

\bibitem[{Moryossef et~al.(2019)Moryossef, Goldberg, and
  Dagan}]{moryossef2019step}
Amit Moryossef, Yoav Goldberg, and Ido Dagan. 2019.
\newblock \href {https://doi.org/10.18653/v1/N19-1236} {{{S}tep-by-Step:
  {S}eparating Planning from Realization in Neural Data-to-Text Generation}}.
\newblock In \emph{Proceedings of the 2019 Conference of the North {A}merican
  Chapter of the Association for Computational Linguistics: Human Language
  Technologies, Volume 1 (Long and Short Papers)}, pages 2267--2277,
  Minneapolis, Minnesota. Association for Computational Linguistics.

\bibitem[{Ng et~al.(2014)Ng, Wu, Briscoe, Hadiwinoto, Susanto, and
  Bryant}]{ng2014conll}
Hwee~Tou Ng, Siew~Mei Wu, Ted Briscoe, Christian Hadiwinoto, Raymond~Hendy
  Susanto, and Christopher Bryant. 2014.
\newblock \href {https://doi.org/10.3115/v1/W14-1701} {{The {C}o{NLL}-2014
  Shared Task on Grammatical Error Correction}}.
\newblock In \emph{Proceedings of the Eighteenth Conference on Computational
  Natural Language Learning: Shared Task}, pages 1--14, Baltimore, Maryland.
  Association for Computational Linguistics.

\bibitem[{Novikova et~al.(2017)Novikova, Du{\v{s}}ek, and
  Rieser}]{novikova2017e2e}
Jekaterina Novikova, Ond{\v{r}}ej Du{\v{s}}ek, and Verena Rieser. 2017.
\newblock \href {https://doi.org/10.18653/v1/W17-5525} {{The {E}2{E} Dataset:
  New Challenges For End-to-End Generation}}.
\newblock In \emph{Proceedings of the 18th Annual {SIG}dial Meeting on
  Discourse and Dialogue}, pages 201--206, Saarbr{\"u}cken, Germany.
  Association for Computational Linguistics.

\bibitem[{Papineni et~al.(2002)Papineni, Roukos, Ward, and
  Zhu}]{papineni-etal-2002-bleu}
Kishore Papineni, Salim Roukos, Todd Ward, and Wei-Jing Zhu. 2002.
\newblock \href {https://doi.org/10.3115/1073083.1073135} {{B}leu: a method for
  automatic evaluation of machine translation}.
\newblock In \emph{Proceedings of the 40th Annual Meeting of the Association
  for Computational Linguistics}, pages 311--318, Philadelphia, Pennsylvania,
  USA. Association for Computational Linguistics.

\bibitem[{Peng et~al.(2019)Peng, Parikh, Faruqui, Dhingra, and
  Das}]{peng2019text}
Hao Peng, Ankur Parikh, Manaal Faruqui, Bhuwan Dhingra, and Dipanjan Das. 2019.
\newblock \href {https://doi.org/10.18653/v1/N19-1263} {{Text Generation with
  Exemplar-based Adaptive Decoding}}.
\newblock In \emph{Proceedings of the 2019 Conference of the North {A}merican
  Chapter of the Association for Computational Linguistics: Human Language
  Technologies, Volume 1 (Long and Short Papers)}, pages 2555--2565,
  Minneapolis, Minnesota. Association for Computational Linguistics.

\bibitem[{Puduppully et~al.(2019)Puduppully, Dong, and
  Lapata}]{puduppully2019data}
Ratish Puduppully, Li~Dong, and Mirella Lapata. 2019.
\newblock \href {https://doi.org/10.18653/v1/P19-1195} {{"Data-to-text
  Generation with Entity Modeling"}}.
\newblock In \emph{Proceedings of the 57th Annual Meeting of the Association
  for Computational Linguistics}, pages 2023--2035, Florence, Italy.
  Association for Computational Linguistics.

\bibitem[{Reddi et~al.(2018)Reddi, Kale, and Kumar}]{reddi2018convergence}
Sashank~J Reddi, Satyen Kale, and Sanjiv Kumar. 2018.
\newblock \href {https://arxiv.org/abs/1904.09237} {{On the convergence of adam
  and beyond}}.
\newblock In \emph{International Conference on Learning Representations}.

\bibitem[{Reed and De~Freitas(2016)}]{reed2016neural}
Scott Reed and Nando De~Freitas. 2016.
\newblock \href {https://arxiv.org/abs/1511.06279} {{Neural
  Programmer-Interpreters}}.
\newblock In \emph{International Conference on Learning Representations}.

\bibitem[{Reiter and Dale(2000)}]{reiter2000building}
Ehud Reiter and Robert Dale. 2000.
\newblock \href {https://doi.org/10.1017/CBO9780511519857} {\emph{{Building
  Natural Language Generation Systems}}}.
\newblock Studies in Natural Language Processing. Cambridge University Press.

\bibitem[{Shen et~al.(2017)Shen, Lei, Barzilay, and Jaakkola}]{shen2017style}
Tianxiao Shen, Tao Lei, Regina Barzilay, and Tommi Jaakkola. 2017.
\newblock \href
  {http://papers.nips.cc/paper/7259-style-transfer-from-non-parallel-text-by-cross-alignment.pdf}
  {{Style Transfer from Non-Parallel Text by Cross-Alignment}}.
\newblock In \emph{Advances in Neural Information Processing Systems 30}, pages
  6830--6841. Curran Associates, Inc.

\bibitem[{Simard et~al.(2007)Simard, Goutte, and
  Isabelle}]{simard2007statistical}
Michel Simard, Cyril Goutte, and Pierre Isabelle. 2007.
\newblock \href {https://www.aclweb.org/anthology/N07-1064} {{Statistical
  Phrase-Based Post-Editing}}.
\newblock In \emph{Human Language Technologies 2007: The Conference of the
  North {A}merican Chapter of the Association for Computational Linguistics;
  Proceedings of the Main Conference}, pages 508--515, Rochester, New York.
  Association for Computational Linguistics.

\bibitem[{Sutskever et~al.(2014)Sutskever, Vinyals, and
  Le}]{sutskever2014sequence}
Ilya Sutskever, Oriol Vinyals, and Quoc~V Le. 2014.
\newblock \href
  {http://papers.nips.cc/paper/5346-sequence-to-sequence-learning-with-neural-networks.pdf}
  {{Sequence to Sequence Learning with Neural Networks}}.
\newblock In \emph{Advances in Neural Information Processing Systems 27}, pages
  3104--3112. Curran Associates, Inc.

\bibitem[{Vaswani et~al.(2017)Vaswani, Shazeer, Parmar, Uszkoreit, Jones,
  Gomez, Kaiser, and Polosukhin}]{vaswani2017attention}
Ashish Vaswani, Noam Shazeer, Niki Parmar, Jakob Uszkoreit, Llion Jones,
  Aidan~N Gomez, \L~ukasz Kaiser, and Illia Polosukhin. 2017.
\newblock \href
  {http://papers.nips.cc/paper/7181-attention-is-all-you-need.pdf} {{Attention
  is All you Need}}.
\newblock In \emph{Advances in Neural Information Processing Systems 30}, pages
  5998--6008. Curran Associates, Inc.

\bibitem[{Vu and Haffari(2018)}]{vu2018automatic}
Thuy-Trang Vu and Gholamreza Haffari. 2018.
\newblock \href {https://doi.org/10.18653/v1/D18-1341} {{Automatic Post-Editing
  of Machine Translation: A Neural Programmer-Interpreter Approach}}.
\newblock In \emph{Proceedings of the 2018 Conference on Empirical Methods in
  Natural Language Processing}, pages 3048--3053, Brussels, Belgium.
  Association for Computational Linguistics.

\bibitem[{Watanabe and Sumita(2015)}]{watanabe-sumita-2015-transition}
Taro Watanabe and Eiichiro Sumita. 2015.
\newblock \href {https://doi.org/10.3115/v1/P15-1113} {Transition-based neural
  constituent parsing}.
\newblock In \emph{Proceedings of the 53rd Annual Meeting of the Association
  for Computational Linguistics and the 7th International Joint Conference on
  Natural Language Processing (Volume 1: Long Papers)}, pages 1169--1179,
  Beijing, China. Association for Computational Linguistics.

\bibitem[{Wiseman et~al.(2017)Wiseman, Shieber, and
  Rush}]{wiseman2017challenges}
Sam Wiseman, Stuart Shieber, and Alexander Rush. 2017.
\newblock \href {https://doi.org/10.18653/v1/D17-1239} {{"Challenges in
  Data-to-Document Generation"}}.
\newblock In \emph{Proceedings of the 2017 Conference on Empirical Methods in
  Natural Language Processing}, pages 2253--2263, Copenhagen, Denmark.
  Association for Computational Linguistics.

\bibitem[{Wubben et~al.(2012)Wubben, van~den Bosch, and
  Krahmer}]{wubben-etal-2012-sentence}
Sander Wubben, Antal van~den Bosch, and Emiel Krahmer. 2012.
\newblock \href {https://www.aclweb.org/anthology/P12-1107} {Sentence
  simplification by monolingual machine translation}.
\newblock In \emph{Proceedings of the 50th Annual Meeting of the Association
  for Computational Linguistics (Volume 1: Long Papers)}, pages 1015--1024,
  Jeju Island, Korea. Association for Computational Linguistics.

\bibitem[{Xu et~al.(2016)Xu, Napoles, Pavlick, Chen, and
  Callison-Burch}]{xu2016optimizing}
Wei Xu, Courtney Napoles, Ellie Pavlick, Quanze Chen, and Chris Callison-Burch.
  2016.
\newblock \href {https://doi.org/10.1162/tacl_a_00107} {{Optimizing Statistical
  Machine Translation for Text Simplification}}.
\newblock \emph{Transactions of the Association for Computational Linguistics},
  4:401--415.

\bibitem[{Yang et~al.(2017)Yang, Halfaker, Kraut, and
  Hovy}]{yang2017identifying}
Diyi Yang, Aaron Halfaker, Robert Kraut, and Eduard Hovy. 2017.
\newblock \href {https://doi.org/10.18653/v1/D17-1213} {{Identifying Semantic
  Edit Intentions from Revisions in {W}ikipedia}}.
\newblock In \emph{Proceedings of the 2017 Conference on Empirical Methods in
  Natural Language Processing}, pages 2000--2010, Copenhagen, Denmark.
  Association for Computational Linguistics.

\bibitem[{Yin et~al.(2019)Yin, Neubig, Allamanis, Brockschmidt, and
  Gaunt}]{yin2019learning}
Pengcheng Yin, Graham Neubig, Miltiadis Allamanis, Marc Brockschmidt, and
  Alexander~L Gaunt. 2019.
\newblock \href {https://openreview.net/forum?id=BJl6AjC5F7} {{Learning to
  Represent Edits}}.
\newblock In \emph{International Conference on Learning Representations}.

\bibitem[{Zhao et~al.(2018)Zhao, Meng, He, Saptono, and
  Parmanto}]{zhao2018integrating}
Sanqiang Zhao, Rui Meng, Daqing He, Andi Saptono, and Bambang Parmanto. 2018.
\newblock \href {https://doi.org/10.18653/v1/D18-1355} {{Integrating
  Transformer and Paraphrase Rules for Sentence Simplification}}.
\newblock In \emph{Proceedings of the 2018 Conference on Empirical Methods in
  Natural Language Processing}, pages 3164--3173, Brussels, Belgium.
  Association for Computational Linguistics.

\bibitem[{Zhao et~al.(2019)Zhao, Wang, Shen, Jia, and Liu}]{zhao2019improving}
Wei Zhao, Liang Wang, Kewei Shen, Ruoyu Jia, and Jingming Liu. 2019.
\newblock \href {https://doi.org/10.18653/v1/N19-1014} {{"Improving Grammatical
  Error Correction via Pre-Training a Copy-Augmented Architecture with
  Unlabeled Data"}}.
\newblock In \emph{Proceedings of the 2019 Conference of the North {A}merican
  Chapter of the Association for Computational Linguistics: Human Language
  Technologies, Volume 1 (Long and Short Papers)}, pages 156--165, Minneapolis,
  Minnesota. Association for Computational Linguistics.

\end{thebibliography}
\bibliographystyle{acl_natbib}

\end{document}